\documentclass[twoside,11pt]{article}

\usepackage{dnd}
\usepackage{times}
\usepackage{url}

\newcommand{\sds}[0]{\textsc{sds}}
\newcommand{\hri}[0]{\textsc{hri}}
\newcommand{\nlu}[0]{\textsc{nlu}}
\newcommand{\nlg}[0]{\textsc{nlg}}
\newcommand{\asr}[0]{\textsc{asr}}
\newcommand{\dm}[0]{\textsc{dm}}

\newcommand{\tts}[0]{\textsc{tts}}
\newcommand{\iu}[0]{\textsc{iu}}

\firstpageno{1}

\begin{document}

\title{Prior Lessons of Incremental Dialogue and Robot Action Management for the Age of Language Models}

\author{Casey Kennington \email caseykennington@boisestate.edu \\ Department of Computer Science \\ Boise State University
      \AND
      \name Pierre Lison \email plison@nr.no \\ Norwegian Computing Center
      \AND 
      \name David Schlangen  \email david.schlangen@uni-potsdam.de \\ Computational Linguistics \\
      University of Potsdam}

\editor{Name Surname}
\submitted{MM/YYYY}{MM/YYYY}{MM/YYYY}

\maketitle

\begin{abstract}%
Efforts towards endowing robots with the ability to speak have benefited from recent advancements in natural language processing, in particular large language models. However, current language models are not fully incremental, as their processing is inherently monotonic and thus lack the ability to revise their interpretations or output in light of newer observations. This monotonicity has important implications for the development of dialogue systems for human--robot interaction.  In this paper, we review the literature on interactive systems that operate incrementally (i.e., at the word level or below it). We motivate the need for incremental systems, survey incremental modeling of important aspects of dialogue like speech recognition and language generation. Primary focus is on the part of the system that makes decisions, known as the dialogue manager. We find that there is very little research on incremental dialogue management, offer some requirements for practical incremental dialogue management, and the implications of incremental dialogue for embodied, robotic platforms in the age of large language models.
\end{abstract}




\section{Introduction}

Large Language Models (LLMs) have become more prominent in robotics, and for good reason. \cite{Williams2024-ng} explain that LLMs can offer ``quick-enabling of full-pipeline solutions" for many aspects of robots ranging from enabling robots to engage humans in spoken interaction to generating action plans \citep{Singh2023-ea,Cohen2024-nx,Singh2024-bi,Mahadevan2024-nz} and emotional behaviors \citep{Mishra2023-wp}. While promising, some recent work has identified important qualities that LLMs lack which, if part of the model, would make interaction with robots seem more natural. A recent survey of spoken interaction on robots by \citet{Reimann2023-ui} showcases a long history of research that \textit{spoken dialogue systems} (\sds s) are key to endowing robots with handling common artifacts in spoken interaction which are not commonly found in text or written interaction such as turn-taking, requests for clarification, building common ground and mutual understanding. At the heart of their focus is the \emph{dialogue manager} because both \sds{}s and robots must \emph{make decisions} about which actions they will take at any given moment, either by uttering a response, moving a robotic arm, or any other potential action within the capabilities of the robot. The \dm\ (or corresponding robot action manager) not only decides which action to take, but also \textit{when} to take that action; both are critical for natural interaction between robots and humans \citep{Lison2023-mu}. A recent review by \citet{Reimann2024-bk} looked at \dm\ in \hri\ tasks and settings comparing how different systems divide the decision-making responsibilities, concluding that \dm\ on robots is still rather a new field; more data, tasks, benchmarks, and discussions are needed. 


Fortunately, there exists a body of literature that spans over 25 years \citep{Allen2001-bn} of success in developing and improving systems that enable humans to talk naturally to machines: \textit{incremental dialogue}, meaning that processing happens at a fine-grained, word-by-word level. Comparisons between incremental and non-incremental systems have shown that incremental systems significantly improve system performance \citep{Ghigi2014-ca}, are perceived by humans as being more natural \citep{Aist2007-mk,Asri2014-yr} and human-like and natural \citep{Edlund2008-ji}, which suggests that the most appropriate systems for robots should be incremental, echoing the requirements of ``robot-ready" \sds\ for use in \textit{human-robot interaction} (\hri) settings \citep{Kennington2020-qx}. 

LLM research can greatly benefit from this knowledge on incremental processing. \citet{Inoue2024-rd} points out that, for example, when humans engage in real-time dialogue with robotic agents, humans expect the robot to take seamless turns (i.e., without a gap in conversation, as happens within human-human dialogue) and they expect backchannels (e.g., nodding, or utterances like \emph{yeah}, or \emph{uh huh}), neither of which are handled with common LLMs. Their work applied \textit{Voice Activity Projection} (VAP) to enable LLMs to predict when a person might stop speaking so the LLM can respond at an appropriate time. \citet{Chiba2025-mr}'s recent VAP method also used an LLM to predict when to start speaking, and their work directly relied on incremental processing. According to the authors: ``[...]even if systems are equipped with a natural turn-taking model, such a model will be ineffective if response generation cannot begin immediately once a turn-shift is detected. Incremental response generation is an approach that addresses this issue." Using LLMs in incremental settings is a positive step, but more work is needed. Furthermore, other recent work \cite{Hudecek2023-aj} asked if LLMs are all that is necessary for task-oriented dialogue (albeit outside of dialogue with robots) with some negatives (e.g., ``LLMs underperform[...]" in important aspects of dialogue) and positives (e.g., ``LLMs show the ability to guide the dialogue to a successful ending"). Finally, \citet{Wagner2024-ov} investigated the usefulness of LLMs in dialogue interaction and found that they are effective, but need control and guidance to ensure that dialogue responses are coherent. 

Taken together, while LLMs can be employed for a wide range of dialogue processing tasks, they still suffer from a number of limitations when it comes to incrementality. In particular, while a LLM decoder can process any kind of input including word-level input, they are trained to act upon complete, sentence-level input, rendering them unable to produce behavior where input and output happen concurrently (an exception in recent full-duplex models shows potential \citep{Zhang2025-jv}). In contrast, humans must see and process individual words while reading text, and psycholinguistic research has shown that speech comprehension happens at a word or even sub-word level \citep{Tanenhaus1995-rb}. Moreover, another requirement of incremental processing is \textit{non-monotonicity}; i.e., that a model can react to change in input, for example when information coming from a speech recognizer is incorrect, the model needs to be able to revoke the erroneous input and change its internal state--causal language modeling is strictly monotonic, but incremental processing should allow for non-monotonic input. 

With chatbots, the text in, text out nature of the interaction is well-suited for LLMs, but the expectation of human-like conversation becomes more challenging when people interact with robots, because robotic interfaces have anthropomorphic characteristics. If, for example, a robot has what appear to be eyes, people expect that the robot can see them, or if the robot has an arm they expect the robot to be able to point or grasp objects. Furthermore, it has been shown that people antrhopomorphize robots for gender \citep{Reich-Stiebert_undated-sl,Eyssel2012-fp}, intelligence \citep{Novikova2015-yx}, and even age \citep{Plane2018-wh} depending on the robot's morphology, size, and movements, which affects the expectations of how robots behave: the more anthropomorphic a robot appears, the more human-like people tend to expect the robot to act. 

In this paper, we review the literature for incremental \sds\ with a particular focus on the decision-making component known as \emph{dialogue management} (\dm; explained further below) for the sake of guiding ongoing and future work related to decision making on robots that interact with humans. We find in our review that ample work has been done in incrementalizing other aspects of \sds{}s such as automatic speech recognition and natural language generation, but there is little work on incremental decision making. We identify some of the challenges and requirements to help guide future research on incremental decision making. The next section begins with background on incremental \sds{}s, focusing first on common modules then fully implemented and evaluated systems. The section that follows then focuses on \dm, giving first a brief overview of \dm\ research, then focuses on incremental \dm. We then end this review with some concluding remarks and suggested paths for future work.


\section{Background: Incremental Spoken Dialogue Systems}

In this section, we review literature on common incremental spoken dialogue system modules except \dm, which we save for the following section. We explain incremental frameworks that have been adopted, and explain different paradigms of modeling incremental processing. 

\subsection{Spoken Dialogue Systems: Overview}

Equally important to the distinction between incremental (word-level) and non-incremental (utterance/sentence-level) \sds\ is the distinction between end-to-end and modular \sds{}s. An end-to-end system is modeled using a single model that takes in input and produces an expected output directly, such as a question-answering system that produces an answer given a question, or social chatbot that produces responses given text input. End-to-end systems often focus on the capability of producing a written or spoken response no matter what the input is. End-to-end architectures now constitute the dominant approaches for developing open-domain dialogue systems where the main focus is the social aspect of the interaction \citep{roller2020open,ni2023recent}. The social aspects of interaction are, of course, important in a natural dialogue, but in task-oriented dialogue there is often something that is required outside of the dialogue itself for the dialogue to be considered successful; e.g., look up information in a database, perform some kind of robotic action, or complete a payment. Modular \sds s  are often \emph{task-based} in that they help the user achieve a goal such as booking a flight; they do not usually focus on social aspects beyond what helps to accomplish the task \citep{budzianowski2018multiwoz,zhang2020recent}. This traditional distinction between open-ended end-to-end systems and task-based modular architectures is, however, increasingly blurry, as recent years have seen the emergence of end-to-end models specifically designed for task completion \citep{liu2018dialogue,zhang2020probabilistic,hosseini2020simple,young2022fusing} as well as newer agentic architectures. Interestingly, end-to-end models for task-oriented systems often operate by augmenting the generative model with implicit ``modules'' in the form of retrieval mechanisms \citep{qin2019entity}, knowledge bases \citep{yang2020graphdialog} or domain-specific ontologies \citep{10.1162/tacl_a_00534}, or by pre-training the response generation model in a modular fashion \citep{10043710}. 

\begin{figure}
    \centering
    \includegraphics[width=0.5\textwidth]{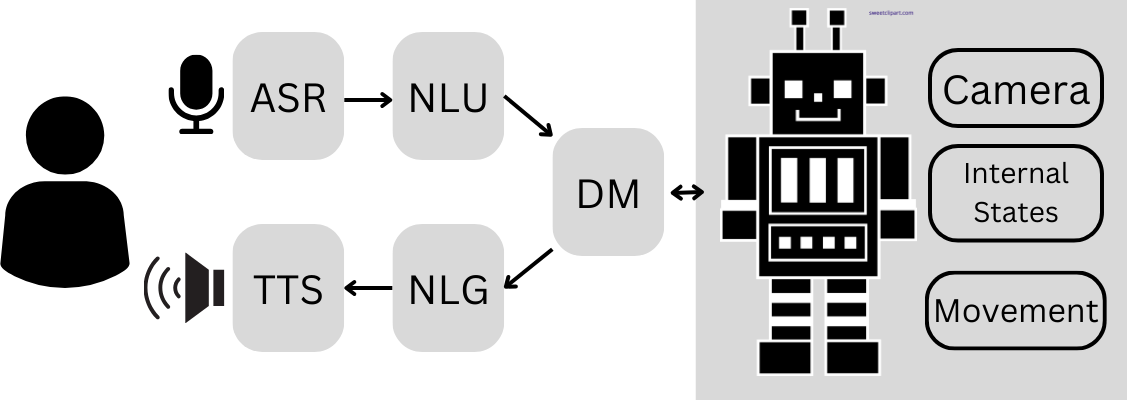}
    \caption{Traditional architecture for spoken dialogue systems composed of Automatic Speech Recognition (ASR), Natural Langauge Understanding (NLU), Dialogue Management (DM), Natural Language Generation (NLG), and Text-to-Speech Synthesis (TTS).}
    \label{fig:sds_examples}
\end{figure}

As the name suggests, modular systems are made up of modules that have well-defined roles in the system, and which can be made to communicate with each other. Figure~\ref{fig:sds_examples} depicts visually a modular \sds. For example, a prototypical \sds\ is often made up five modules including automatic speech recognition (\asr{}) that transcribes speech to a text representation of the human utterances, natural language understanding (\nlu) that takes the text and yields a computable semantic abstraction, dialogue management (including dialogue state tracking) that makes a high-level decision about the next action to take (e.g., look up information in a database and respond to the user), natural language generation (\nlg) that takes the dialogue manager's decision and determines which words to use and in what order, and text-to-speech (\tts) which actually speaks the words. These modules are further explained below. 


Modular \sds{}s that process incrementally have an added complexity in that all of the modules must operate at granularities that downstream modules can make use of, such as at the word level from \asr\ to \nlu. For example, given a system on a robot that is made up of the standard five modules as explained above (along with connections to robotic modules), and someone utters \emph{Hand me the green book on the left}, an incremental \sds\ begins to process as soon as speech is detected. The \asr{} \ outputs each word, one at a time, and the \nlu\ updates its understanding state each time a word is outputted by the \asr{}, and the \nlu\ likewise produces outputs as it gathers information about the utterance, for example, tagging \emph{hand} as the action as the first word is uttered is uttered, and a specific book as the target once \emph{the green book} it has been uttered. The \dm\ is tasked with querying a module that takes in visual information and instructing an arm to reach for the book in question. An incremental \dm\ might already extend its arm in no particular direction as the first word is uttered, then towards any green book once \emph{green book} is uttered, then narrow the target down further as \emph{on the left} is uttered. The \nlg\ actually then could utter something like \emph{green book} as it begins to move its arm then \emph{ah, here we go} once it determines a unique referent. 

The above example highlights some things that differentiate an incremental \dm\ from a more traditional \dm. First, the incremental \dm\ receives installments of information over time, whereas a traditional \dm\ receives all of the information at once after everything has been uttered and the \nlu\ processes the \asr's transcription. The incremental \dm\ has an important job to do that the non-incremental \dm\ does not: it not only must decide which action to take, but it also must decide \emph{when} to take that action given the information that it has so far, and---perhaps a bigger challenge---perform concurrent actions as it is still receiving input (i.e., ``full-duplex" models). Traditional \sds\ has often relied on endpointing; i.e., waiting for silence after a speaker begins to speak, which burdens the \asr{} \ with determining \emph{when} to act. However, pauses in speech are not always signals that someone is done speaking, and incremental \sds\ that relies on a \dm\ to determine when to take an action can potentially use speech, silence, as well as information from the content of the utterance (i.e., via the \nlu) to make decisions about \emph{when} to act.

\subsection{Frameworks \& Architectures}

\subsubsection{The Incremental Unit Framework}

A well-known framework for incremental processing that we will make reference to throughout this paper is the \emph{incremental unit} (\iu) framework \citep{Schlangen2009-ob,Schlangen2011-wi}. The \iu\ framework views each bit of information created by the modules (e.g., words produced by \asr{} \ and slots produced by \nlu) as part of a global network of interconnected \iu s no matter which module produced them. The framework defines functions for changing the network including how nodes of the network are added and how the nodes are interconnected. Newly created \iu s by a module (e.g., words by \asr{} ) can be \emph{added} to the \iu\ network, \emph{revoked} from the network if the module determines that an \iu\ was erroneously added in light of new information (e.g., the \asr{} \ first added the word \iu\ \emph{four} but later revoked and added \emph{forty}), and \iu s can be \emph{committed}, meaning they have already been added to the \iu\ network, and are guaranteed to not be revoked. 

To be added to the \iu\ network, an \iu\ has to be connected to other \iu s that already exist in the network through two relations: \emph{same level links} which are relations between \iu s created by the same module (e.g., if the \asr{} \ recognizes \emph{the} and \emph{dog} as two \iu s, the later word \emph{dog} as a same level link to \emph{the}), and \emph{grounded-in links} where a relation is created between and \iu\ an the \iu(s) that gave rise to that \iu, for example the \iu s \emph{the} and \emph{dog} from the \asr{} \ might give rise to a \emph{subject} tag in the \nlu, so it will have to grounded-in links, one to each word \iu. When \iu s are operated on (i.e., added, revoked, or committed) the modules that triggered the operation signal downstream modules that consume their input about the change. For example, as the \asr{} \ module recognizes words from a microphone, it adds each of them to the \iu\ network and signals to the \nlu\ module that a new word has been added. 

A module's ability to revoke is very important in a natural dialogue interaction. The above \asr\ example of revoking is fairly straight-forward and happens internally between the modules, but there are cases where modules that produce output need to revoke information, for example a robot is planning on uttering something that, given new information, should be changed. In a setting where a robot and a person are working together to move around objects including many boxes in many different colors, and the person says \emph{move the green box to the left} and the \asr\ originally recognized \textit{gray} instead of \textit{green}, then the robot will first move towards and attempt to move a gray box. The robot generates a plan to move towards the box and produce an utterance to signal understanding such as \textit{okay, I'm on my way to move the gray box}. But when the revoke from \textit{gray} to \textit{green} happens, the robot needs to adjust its plan in where it is moving and what it is saying. If the robot hasn't completed its utterance, it still has a chance to change the utterance to \textit{okay, I'm on my way to move the green box} and change its direction to the box the person referred to, known in robotics as \textit{replanning} \citep{Cashmore2019-pk}. This is an illustration of \textit{non-monotonicity}: modules can update the information that they pass to each other, and often updates need to happen after the robot has already taken some kind of action. More on this in Section~\ref{sec:idm}. 

The \iu\ framework has been implemented in several software packages, notably in Java as InproTK \citep{Baumann2012-xh} and more recently in Python as Retico \citep{Michael2019-dx} and Remdis \citep{Chiba2024-zw}. Other conceptual frameworks such as the Information State Approach \citep{Traum2003-zl} and Cohen's belief-desire-intent model \citep{Cohen2017_bdi} remain valid in incremental \sds, including within the \iu\ framework, though they are not strictly incremental dialogue frameworks. Later versions of InproTK and, more recently, Retico has been extended to include common robot capabilities such as object detection and control of multiple robot platforms \citep{Kennington2020-qx}, towards bridging the gap between \sds\ and \hri\ research. While not strictly incremental, OpenDial (Java and Python) has been used in spoken \hri\ studies, and can serve as a decision making module in both InproTK and Retico \citep{Lison2016-jk,Jang2020-fb}. 

\subsubsection{Restart vs. Update Incremental Models}

\citet{Khouzaimi2014-vf} points out that not all methods and models are inherently incremental, though many can be made to work incrementally under certain constraints. While their proposed method is important step for making systems incremental, we point out here that there are two ways to approach modeling incremental systems which has implications for how an incremental \dm\ could work: restart incremental and update incremental, which we explain presently.

\paragraph{Restart Incremental} Restart incremental models can take in input and produce incremental output (e.g., at the word level), but the input is repeated as the prefix grows, and models themselves are agnostic to the incremental updates. Any model (e.g., a language model using zero-shot classification) could be used restart incrementally. For example, a \nlu\ module that is restart incremental would take in the following input (time moves from top to bottom; each line represents input to a \nlu{} model): \\

    \begin{tabular}{ccc}
        the &  &  \\
        the & dog &\\
        the & dog & barks \\
    \end{tabular}

\paragraph{Update Incremental} In contrast to restart incremental models, update incremental models do not need repeated input and the model is designed to maintain a state that updates for each incremental input. An \nlu\ model that works in an update incremental way would not need to repeat a growing prefix from the \asr{} : \\

    \begin{tabular}{l}
        the  \\
        dog \\
        barks \\
    \end{tabular}
\\ \\
The model explained in \citet{Kennington2017-jn}, for example, is a Bayesian model that produced a distribution over possible slot values that updated the distribution at each word increment. An open question that we explore below is if a \dm\ model should be either restart or update incremental.

\subsection{Common Modules in Incremental, Interactive Systems}

\subsubsection{Automatic Speech Recognition} Current \asr\ systems receive streaming input and produce partial transcriptions, and can often work at word-level increments. Early incremental \asr{} \ was implemented in Sphinx \citep{Baumann2009-de}, and recent, neural \asr\ systems can often operate at the character level \citep{Hwang2016-pa}. The most common evaluation of \asr\ is \emph{word error rate}, and recent neural models are showing very low error rates in common \asr\ benchmark datasets. However, evaluation of incremental \asr\ requires a closer look at how often a model alters its output and latency of results \citep{Baumann2016-ao,Whetten2023-vu}. Because of the nature of the input and incremental output of \asr, it can be evaluated in isolation though it is also important to evaluate \asr\ in larger systems because certain mistakes will propagate to downstream modules. 

Because \asr{} requires streaming input, they are inherently incremental in terms of input, though not all recognizers produce incremental output--they often wait until a pause in the speech (i.e., end-pointing). However, recent \asr{} models have become very effective at accurate transcription in multiple languages and they produce incremental output. The Whisper \citep{Radford2022-fn}, wav2vec \citep{Baevski2020-zw}, and Deep Speech 2 \citep{Amodei2016-pu} all produce incremental output, the former 2 have been incorporated into the Retico framework. \citet{Imai2025-yo} recently evaluated conversational speech recognition on several state-of-the-art \asr\ models (including Whisper and wacv2vec), taking gender into account in their evaluation. The results were mixed; \asr\ has come a long way in the past two decades, but more work needs to be done to accommodate different demographics of speakers, and correcting errors. 

Incremental \asr\ is made more challenging on robots in spoken \hri\ settings because robot voice can be picked up by the \asr, thereby 'confusing' the robot. This can be at least partially addressed using diarization (i.e., tracking the voice of particular individuals within a speech signal, including a robot) or by the robot tracking its own speech signal and filtering it out of the \asr\ input. 

\subsubsection{Natural Language Understanding} Understanding natural language in \sds{}s also has a long history. In most \nlu\ models, the input is text. It the case of \sds, the input to \nlu\ is transcribed speech. The output of \nlu\ is important to consider here, because it is often what serves as the input to the \dm. The output needs to abstracted sufficiently over the input text to form a computable meaning representation that the \dm\ can use for making a decision on how to act. That meaning representation in incremental \nlu\ has been represented in various ways in the literature including tagged words, logical forms, or frames (i.e., a set of key-value pairs known as \emph{slots}), recent models tend to use tags to produce slots and frames as output, or latent representations (e.g., embeddings). Below is an example frame for an utterance made to command a specific robot action: \emph{Move the red ball into the box on the left} made up of four slots: \\

\begin{center}

    \begin{tabular}{ll}
        \texttt{intent} & command \\
        \texttt{object} & red ball \\
        \texttt{target} & left box \\
        \texttt{action} & move \texttt{object} to \texttt{target} \\
    \end{tabular}
\end{center}

Like incremental \asr{}, incremental \nlu\ produces output as much as possible as early as possible (for example, individual filled slots), but unlike \asr{}, the input is discrete words instead of a continuous speech signal, so the intervals of when output is produced can vary depending on the input and the domain. Early incremental \nlu\ focused on classifying frames. Each input word produced a partially complete frame as output \citep{Devault2011-oi,DeVault2012-gs,DeVault2013-it,Yamauchi2013-ok,Kennington2014-sp,Kennington2014-co,Kennington2015-el}. Part of the frame is also the dialogue act; i.e., the overarching type of utterance produced by the interlocutor (e.g., a question or an assertion), which has also a history of incremental models \citep{Petukhova2011-vi}. Early actionable output from \nlu\ is particularly important in \hri\ settings, where robots can already be moving towards referred objects before the person finishes their request, which is an important physical backchannel: a robot beginning to move towards an object is a signal to the user that the robot is understanding the unfolding utterance, as done in \citet{Hough2016-oo}

Similar to their non-incremental counterparts, incremental \nlu\ can benefit from syntactic parsing to guide the language understanding, but in the case of incremental \nlu, the parsers must also work incrementally (i.e., produce a partial syntactic parse such as a tree for each word input). There has been ample research in incremental parsing for different syntactic theories, including dependency parsing \citep{Nivre2008-ft}, combinatory categorical grammar parsing \citep{Hassan2008-sz,Beuck2013-vu}, as well as formalisms that have more semantic relational information including  robust minimal recursion semantics parsing \citep{Copestake2007-xp,Peldszus2012-jh}, dynamic syntax \citep{Eshghi2013-lm} (see \citet{Hough2015-id} for a comparison of robust minimal recursion semantics and dynamic syntax for incremental dialogue), as well as abstract meaning representation \citep{Damonte2017-px}. 

In multimodal \sds{} and interaction with robots, there is sometimes a need for the \nlu\ to resolve references to objects that exist in the shared space with the robotic system and the human interlocutor. Incremental reference resolution can be viewed as the ability to narrow down possible referents in a shared visual space to an individual object. An incremental reference resolution model might, for example, understand the word \emph{red} to refer to objects that have a red color, and \emph{book} to then further narrow down from all red objects to only red books. Incremental reference resolution is sometimes an integral part of \nlu\ \citep{Kennington2014-co}, but have also been designed for modules that only resolve references \citep{Schlangen2009-bf,Paetzel2015-iv,Kennington2015-ia,Schlangen2016-mz,Kennington2017-jn}, information that the \dm\ may need to use for making a decision. 

Other work has explored how deep learning architectures can be used for incremental \nlu, including recurrent architectures \citep{Shivakumar2019-vd} and to what degree architectures that are not inherently incremental (e.g., self-attention transformers which are designed to process multi-word input in parallel) can be used for incremental \nlu\ \citep{Madureira2020-fr}, with mixed results. It is important to explore further how neural models can work incrementally because many dialogue phenomena are incremental in nature. For example, \citet{Shalyminov2017-wq} showed that that deep neural dialogue models failed on common spoken phenomena like restarts and self-corrections. 

More recent work has explored how transformer LMs can be successfully used for incremental NLU. \citet{Madureira2020-fr} showed that both bidirectional Long short-term memory (LSTM) recurrent models and transformer-based encoders assume that an input sequence to be encoded is available a-prior in its entirety, to be processed either forwards and backwards (in the case of bi-directional LSTMS) or as a full sequence (in the case of transformer-based encoders). The results of their work support the possibility of using bidirectional encoders in their developed \textit{incremental} mode while training their non-incremental qualities (i.e., parallel processing). \citet{Kahardipraja2021-fv} explored using \textit{linear} transformers with a recurrence mechanism to examine the feasibility of linear transformers for incremental NLU. They found that linear transformers have better performance and faster inference than standard transformers when used in a restart-incremental fashion. 

\hri\ settings make the requirements of \nlu\ more challenging due to the fast-paced, multimodal nature of the interaction. LLMs trained only on text are limited, but the recent proliferation of multimodal LLMs have real potential for being used on robot platforms. Modalities include images, speech, and video, for example the ONE-PEACE \citep{Wang2023-fa} and PALM-E models \citep{Driess2023-sf}. A recent survey explains the modeling trends (e.g., one vs. two-tower; different methods of representing images) for vision LMs \citep{Fields2023-zv}; improvements in visual LMs will benefit \hri\ research because robots need to see and talk about objects in a shared space.

Other recent work focuses on how transformers handle incremental NLU revisions. Most transformer models are \textit{causal} in that they are forced to produce a single output once an ambiguity is resolved, but \cite{Madureira2024-io} proposed an interpretable way to analyse incremental states to show how transformers handle ambiguity. They showed that transformer sequential structures encode information on the garden path effect, as well as the resolution of garden paths. Another model, TAPIR, a two-pass method that modeled the revision process itself showed better performance on incremental metrics compared to transformers used restart-incrementally \citep{Kahardipraja2023-mb}. 

\subsubsection{Natural Language Generation and Speech Synthesis}


Early work in incremental \nlg\ focused on resolving references in situated dialog. \citet{Kelleher2006-ib} presented an approach to generating locative expressions using a basic incremental algorithm that considered \emph{visual salience} as a computation of an object's perceivable size and centrality relative to the viewer, choosing words that will distinguish between the target object and distractor objects. While the algorithm the authors present is ``incremental", it is not evaluated as a word-by-word incremental model, but given the co-location and potential of being used at the word level, we include it here. More recent work has shown that incremental installments of words that refer to a visual object using a model trained on visual object/word pairings that uses a beam search to determine the best possible word to utter can use a model of vision/word that isn't trained specifically for \nlg\ \citep{Zarries2016-gb}.

Incremental \nlg\ that builds on the \iu\ framework included work that used a buffer of words to be uttered, and three operations \textsc{add}, \textsc{revoke}, and \textsc{purge} were used for operating on the buffer \citep{Dethlefs2012-dn}. The \textsc{add} operation, of course, means a word is added to the buffer and eventually uttered, unless it was \textsc{revoke}d (removed from the buffer) or \textsc{purge}d (all words currently in the buffer are removed in favor of a new hypothesis/goal). The \nlg\ module often produced words faster than they could be articulated by a \tts, giving an incremental \nlg{} time to determine which words should be uttered, and in which order. The authors also carried out experiments to explore how \nlg\ interacts with output generation of other modalities, such as information on a screen \citep{Dethlefs2012-kk}. In general, the research has shown how incremental generation produces systems that are more reactive and perceived as more natural to human dialogue partners. In \hri\ settings, incremental \nlg\ is important, for example, when referring to real-world objects \citep{Zarries2016-gb}. 

Others also looked at how incremental multimodal generation affects the interaction qualities when the \sds\ is part of a virtual agent; articulation of course included \nlg, but also hand gestures and eye gaze by the agent \citep{Van_Welbergen2012-tp}. Instead of planning all articulations before they were realized, the model generated behaviors incrementally and linked increments in the multiple output modalities to each other so what happened corresponded temporally to other modalities (e.g., saying \emph{that} in conjunction with a pointing gesture). Such articulation means that the speech synthesis must also be incremental because an ongoing utterance that is offered by the \nlg\ to the \tts\ might change before the \tts\ actually articulates a word in the utterance, thereby changing prosody or duration; e.g., the system may want to hold the floor longer so will need to take longer to speak or insert artifacts such as \emph{ummm}  \citep{Buschmeier2012-iu,Baumann2014-ac}.

Improvements in \asr\ and \nlu\ have enabled systems to be far more capable than even a few years ago, but the biggest gains have been in \nlg\ due to LMs. Generative large LMs are flexible in that inputs can be structured text and models can be made to produce useful structured output that is useful for \sds\ and \hri. For example, the \dm's can output a request to look up information in a database, then take that structured information and input it into a LM, which produces a surface utterance that has the necessary information. Fortunately, while LM \textit{input} processing is not incremental as defined, LM \textit{output} is naturally incremental due to inherent modeling (i.e., autoregression). However, while LM output can be paused \citep{Goyal2023-lc} or interrupted, the output itself cannot be changed given new input.

\subsubsection{Incremental Systems \& Evaluation}

\paragraph{Incremental systems improve over non-incremental counterparts} Beyond individual modules, full systems are more complex and difficult to evaluate, but some have shown how incremental systems are better in some domains than non-incremental counterparts. For example, a virtual in-car dialogue that presented information incrementally was shown to be safer and more effective at helping users remember information \citep{Kousidis2014-xd}. The system was able to detect changes in the car's control (e.g., changing lanes or speed) and if any change was detected, the system would pause its output and resume after the driving was constant. This allowed drivers to focus on driving instead of non co-located interlocutors. 

In another system, \citet{Fischer2021-iy} used incremental speech adaptation to initiate human-robot interactions in noisy (in-the-wild) scenarios. The robot incrementally adjusted the loudness of its voice depending on the circumstances, and was perceived positively by human users. Finally, \citet{Ghigi2014-eq} showed that that an incremental dialogue strategy significantly improved system performance by eliminating long and often off-task utterances that generally produce poor speech recognition results. User behavior is also affected; the user tends to shorten utterances after being interrupted by the system.

\paragraph{Challenges of incremental evaluation} \citet{Kohn2018-it} reviewed incremental processing in the field of natural language processing (including parsing, machine translation, among others which are beyond our scope), pointing out that granularity, grounding, monotonicity, and timeliness are all aspects of incremental processing that play into how incremental systems are perceived. Most incremental \sds\ research is performed with the level of granularity set at the word level, but it might be better in certain cases to work at sub-word or phrase levels, or on speech directly (see \citet{Kebe2022-jm} for a non-incremental model that grounds language with raw speech). Grounding, moreover, is how a system aligns its output (in the case of \sds, generated speech) to what is happening in the dialogue state including physical context and the conversation up until that point. Grounding is particularly important (and challenging) for \hri\ settings and tasks, as the human and the robot need to track objects, dialogue history (including entity tracking; i.e., objects that have been discussed before). Task-completion is a common metric (i.e., did the human and robot pair complete the assigned task, like put together a puzzle), but in many cases the task is more social and more focused on human impressions of their interactions with the robot. 

Another challenge in incremental evaluation that is more specific to incremental processing is monotonicity. Monotonicity is an open question in \sds\ research; an ideal incremental \asr{}, for example, would only output the correct word as early as possible as they are spoken without the need for revoking and replacing words. Thus while monotonicity is an ideal to strive for, system modules make mistakes and need to be able to repair those mistakes (hence the need for the \iu\ framework), but knowing how monotonic a system or an individual module is can be a useful metric for measuring stability. Finally, timeliness is important: the system needs to respond quickly, but the system should reach a level of confidence that the response is the proper one. Thus the challenge is balancing these two competing optimizations: timeliness with accuracy, highlighting the need for non-monotonic operations.

\section{Review of Incremental Dialogue Management} \label{sec:idm}

In this section, we review literature relating to incremental \dm. We give an overview of \dm, dialogue state tracking, and attempts at incremental \dm. 

\subsection{A Brief Overview of Dialogue Management}

Dialogue management lies at the crossroads between \nlu{} and \nlg{} and is responsible for controlling the general flow of the interaction, often in relation with the task(s) that should be fulfilled by the dialogue agent. In their seminal work on the \emph{Information State} approach to dialogue management, \citet{Traum2003-zl} mention four objectives:
\begin{enumerate}
    \item updating a representation of the dialogue context on the basis of interpreted communication (from all dialogue participants) ; 
    \item providing context-dependent expectations for interpretation of observed signals as communicative behavior ;
    \item interfacing with task/domain processing (e.g., database, planner, execution module, other back-end system), to coordinate dialogue and non-dialogue behavior and reasoning ;
    \item deciding what content to express next and when to express it.
\end{enumerate}

Current \dm{} approaches distinguish between two central (and consecutive) components, respectively called \emph{dialogue state tracking} and \emph{action/response selection}. 

\subsubsection{Dialogue State Tracking} 

The task of maintaining a representation of the current dialogue state over the course of the interaction is called \emph{dialogue state tracking} \citep{williams2016dialog,ren2018towards,heck2020trippy}. The dialogue state aims to reflect the system knowledge of the current conversational situation, and often includes multiple variables related to the dialogue history, common ground, external context (including the physical context, in the case of human--robot interaction), and the task(s) to perform. 

This update of this dialogue state should occur upon the reception of any new observation that may potentially impact the system's understanding of the current conversational situation, such as new user utterances, but also changes in the physical context of the interaction (for instance, new entities perceived in the visual scene, or updates on the current location of the robot). For incremental systems, those observations will typically correspond to incremental units produced by the \nlu{} module. 

In task-oriented systems, the dialogue state is often represented  as a list of slots to fill \citep{williams2016dialog,mrkvsic2017neural}, where a slot typically represents a required or optional attribute whose value should be derived from the user inputs to complete the task. For instance, a restaurant booking system might have slots for the date, time and number of people. Although such slot-filling representation can be applied to many domains, it remains restricted to a fixed list of predefined slots, and may therefore be difficult to apply to conversational domains with varying numbers of entities and relations between them. This is notably the case in human--robot interaction, where the number of persons in a room, or the number of objects detected in the current visual scene is not fixed in advance and may change over the course of the interaction. In such settings, representing the dialogue state as a \emph{graph} of entities connected through various relations is a preferred alternative \citep{ultes-etal-2018-addressing,walker2022graphwoz}.

Approaches to dialogue state tracking also differ in whether they explicitly represent uncertainty related the current dialogue state using probability distributions. Many dialogue management approaches represent the current dialogue state as a mere collection of key-value pairs (slots and their values). Although this representation does simplify both dialogue state tracking and action selection (in particular when this selection is optimized using reinforcement learning), it makes it harder to capture uncertain, ambiguous or untrustworthy information, which may arise from e.g.~error-prone sensory inputs (e.g.~imperfect object recognition or \asr{}) or non-deterministic inference (e.g.~linguistic ambiguities). An alternative is to explicitly represent the dialogue state as partially observable and define a probability distribution over possible state values \citep{Young2013-ny,mrkvsic2017neural}, often called the \textit{belief state}. This belief state can notably be expressed as a Bayesian network over state variables \citep{thomson2010bayesian}.

\subsubsection{Action/response selection} 

The second core \dm{} task is \emph{action selection}, whose role is to determine the next (verbal or non-verbal) action(s) to undertake by the system, based on the dialogue state updated through dialogue state tracking. Although those actions frequently correspond to verbal system responses, they may also express other types of actions, such as API calls or high-level physical actions in the case of robotic platforms. A given dialogue state may lead to the selection of several actions to execute in parallel or in sequence (for instance, a robot may simultaneously move to a new location and utter a sentence to describe his movement to the user) or to no action at all. 

The selection of the next action/response may take several forms, from handcrafted flowcharts and logical rules to data-driven techniques. Early work includes \citet{Larsson2002-vt}, which surveyed existing approaches to \dm\ including logic-based, finite state, form-based, and plan-based approaches, but the author regarded those approaches as limited in their practicality---most were theoretical models without a concrete implementation. To remedy this situation, \citet{Larsson2002-vt} introduced \textit{Issue-based Dialogue Management}. Issues are modeled semantically as questions, which can be implemented in multiple theories (e.g., plan-based or form-based). This kind of dialogue is system-driven in that the system has a specific task that it must perform and it drives the dialogue by asking questions to the user, for example an automated travel agency would ask questions about price ranges, travel dates, origin and destination airports, and airlines if it is going to help a user find an appropriate flight. As is the case with most dialogues, a kind of ``information exchange" takes place, the system is not requiring anything of the user beyond responding verbally with requested information. 

Also seminal is the early work of \citet{Cohen1990-dk} on plan-based approaches to dialogue management, building on earlier work by \citet{Allen1979-tj}. More recently, \citet{Cohen2023-nw} showcases a fully working multimodal conversational system that infers users' intentions and plans to achieve those goals. The system can infer obstacles to goals and actions and find ways to address them collaboratively. The \dm\ is broken down int four parts: plan recognition, obstacle detection and goal adoption, planning, then execution. Planning here is an important aspect of the \dm; it does not just identify an action to take now, it identifies a plan (i.e., a series of actions) that must be taken to achieve a higher-level user goal, making it potentially more amenable to multimodal (including IVA and robotic) control. 

The mapping from dialogue state to action(s) is called a \emph{dialogue policy}, and various methods have been developed to automatically learn such policies from real or simulated dialogue data. Supervised learning techniques can be employed to imitate the conversational strategies followed by human experts in a corpus of dialogue \citep{griol2008statistical}. However, the behavior of human experts may be hard to imitate, especially as those experts  often base their decisions on a different and richer understanding of the conversational context that what can be captured in a dialogue state. Those supervised learning techniques also suffer from data sparsity problems, as only a small fraction of the state space can realistically be covered by the dialogue examples. 

To this end, a range of reinforcement learning methods have been proposed to automatically optimize dialogue policies based on a reward function \citep{rieser2011reinforcement,young2013pomdp,williams2017hybrid,peng2018deep}. Although the reward function is often defined manually based on the system objectives, it can also be learned from data \citep{su2018reward,takanobu2019guided}. The dialogues can be generated automatically using a user simulator \citep{schatzmann2006survey,chandramohan2011user,ultes2017pydial} or from actual dialogues with human users \citep{su2016line,shah2018bootstrapping}.

The underlying process to optimize may be either framed as a Markov Decision Process (MDP), or, in case the dialogue state itself is consider to be uncertain, a Partially Observable Markov Decision Process (POMDP). While framing action selection as a POMDP makes it possible to explicitly account for uncertainties about the current dialogue state, it also complicates the dialogue policy optimization, due to the need to derive a policy in a continuous and high-dimensional belief state space. Dialogue policies can also be expressed in terms of probabilistic rules with a skeleton provided by the system designer while the rule parameters are estimated from dialogue data, as shown by \citet{Lison2015-ye,lison2015hybrid}. Recent work in reinforcement learning goes well beyond the POMDP model, including reinforcement learning with human feedback \citep{Ouyang2022-tt}, and proximal policy optimization \citep{Shao2024-eh}, each with potential effectiveness for \dm. Some recent work shows how LMs can be used for \dm\ \citep{Niu2024-lo,Zhang2025-jv}. 

Little work has been done, however, on the problem of \emph{revising} current action plans of the dialogue manager in light of new (incremental) observations. For instance, a robot may start executing a particular action plan, but suddenly hear a human user say ``stop!", in which case the robot ought to interrupt its current sequence of actions and devise an alternative response. This ability to revise or regenerate plans is related to the problem of replanning in robotics and automation \citep{Garrett2020-tr,Zhou2023-eg}.

The output of the action selection should convey what the system should say or do next, and is often structured as a logical form \citep{Traum2003-zl,lison2015hybrid}. In the case of a verbal response, the \nlg{} module is then responsible for converting this representation into an actual utterance. Alternatively, the dialogue manager may generate a prompt containing natural language instructions on how to respond, and use this prompt as input to a LLM in charge of producing the response.

\subsubsection{Turn-taking and End-of-turn Prediction}

Dialogue management in spoken dialogue systems is not only about what to do, but also about when to do it. This question of timing has, unfortunately, not received as much attention as it should have.  A common but sub-optimal approach is to wait until the current speaker has stopped speaking for a given period of time, and seek to predict whether they are likely to continue or not \citep{ferrer2002speaker}. \citet{raux2009finite} presented a finite-state model for turn-taking in spoken dialogue systems, relying on a cost matrix and a decision-theoretic framework to determine whether to grab the dialogue floor, release it, wait or keep the floor.  Several machine learning models have also been developed to automatically predict when the utterance of the current speak is about to end \citep{de2009multimodal,maier2017towards}. \citet{roddy2018investigating} presented a data-driven approach to predict a range of turn-taking behaviours when encountering pauses or overlaps, based on on speech-related features. \citet{skantze2021turn} and \cite{Ohagi2024-cq} provide a general survey of the various approaches to turn-taking in both embodied and non-embodied speech-based dialogue systems. 

Early deep learning approaches to end-of-turn prediction include \citet{Maier2017-ax}, which applied a long short-term memory model to the task using live acoustic features. Such recurrent models are inherently incremental. Using transformer LMs to predict turn-taking is well represented in the recent literature. While not inherently incremental as defined above, turn-taking requires models to handle continuous input. TurnGPT made early use of LMs to detect turn shifts in dialogue \citep{Ekstedt2020-xz}, with some discussion as to how the model could be used to predict end-of-turn. This work was extended in \cite{Inoue2024-ne}, which uses VAP which includes contrastive predictive coding of a cross-attention transformer (as a plus, the model is effective on a CPU). More recently, \citep{Chiba2025-mr} also applied VAP within an incremental framework (in their case, Remdis \citep{Chiba2024-zw}) and \cite{Roddy2020-hu} proposed a model of response timing that is designed for use in incremental systems; human evaluations indicated that they perceived the interaction qualities as more natural when the model was in use. \cite{Shukuri2023-nr} were also concerned with timing and turn-taking and proposed a method for using LMs as meta-controllers of dialogue (where the dialogue system is made up of LMs). 

\subsection{Incrementalizing Dialogue Management}

\begin{figure*}
  \center
  \includegraphics[width=1.0\textwidth]{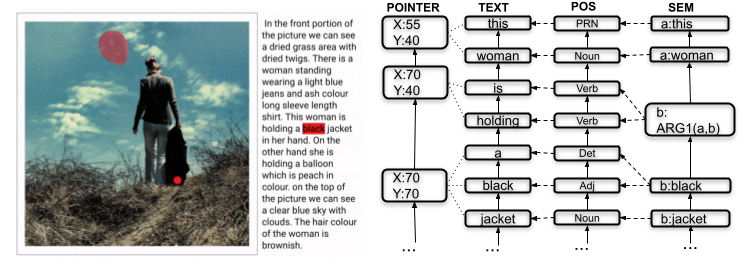}
  \caption{From \citet{Kennington2021-fm}, an example of  Pointer, Word, POS, and SEM \iu\ annotations for a sample from the Localized Narrative dataset. Solid lines denote \textsc{sll}s, dashed denote \textsc{grin}s, and the dotted lines denote an alignment between two modalities. Image taken from https://google.github.io/localized-narratives.}
  \label{fig:locnar-example}
\end{figure*}

\citet{Bus2010-jb} introduced an Information State Approach to incremental \dm\ using the \iu\ framework where the \iu s themselves composed the information state. In their method, they focused on the collaborative nature of many dialogues in a micro domain of playing a puzzle game. All modules, including \asr{}, \nlu, \tts, and a floor tracker were modeled at the incremental word level. The incremental \dm\ reacted to information from the \nlu, game board state change (i.e., non-linguistic relevant state actions), and the floor tracker. The central element of information was the iQUD (incremental QUD, following \citet{Ginzburg2012-wm}) and is rule-based. They evaluated using an incremental and a non-incremental version of their system and found that the incremental versions were rated higher human-likeness and reactivity by human observers of recorded dialogue of both incremental and non-incremental interactions. This is promising, but limited as a methodology for incremental dialogue. 

The same authors followed up this work with \citet{Bus2011-es} that introduced DIUM---dialogue incremental unit manager---that is also rule-based, but builds on their prior work by leveraging edits that can be made in a dialogue system that is built on the \iu-framework. One positive aspect of incremental dialogue is that systems can respond appreciably faster than non-incremental counterparts, but a potential drawback of early response is that the response is based on information which has already been, or is currently being, updated in processing modules. For example, an \asr{} \ recognizes \emph{I would like to book a train to Hamm} passes each word to a \nlu\ module that informs the \dm\ with information about which action to take and which object to take the action on. The \dm\ begins to act by looking up train information in a database and informing the \nlg\ about how to respond, and \tts\ begins to vocalize the response, but at that moment \emph{Hamm} is revoked and replaced with \emph{Hamburg}. This \asr{} update is propagated to the \nlu\ and likewise to the \dm. What action should the \dm\ now take given that \tts\ is currently uttering something about the wrong city? This is a shortcoming of incremental systems that needs to be addressed according to the authors. Instead of reducing revisions (i.e., waiting for more information) which means waiting longer, and instead of ignoring the problem completely, DIUM offers a third alternative: acknowledge the problem and repair it explicitly. The \iu\ information state is adaptable to addressing the problem directly because a revoke---an important part of the \iu-framework---is an abrupt change to the information state that can be addressed by triggering an explicit repair, for example \emph{Oops, I thought you said Hamm, but it was actually Hamburg. Let me get that information for you.}

Unfortunately, this line of research has not been pursued since the 2011 DIUM paper. However, recently, \citet{Kennington2021-fm} proposed a sketch of using the \iu-framework as a method of representing a multimodal, fine-grained information state for use in physical, co-located settings such as \hri. Like \citet{Bus2011-es}, their sketch explained how the information state can consist of the full \iu\ network including connections between \iu s, as well as all prior edits. Figure~\ref{fig:locnar-example} shows an example of a fine-grained, incremental information state using an example from the Localized Narrative Dataset \citep{Pont-Tuset2020-vy}.

Later work explored incremental \dm\ using a \emph{Time Board} where input, output, and decisions made by the DM are posted on the Time Board \citep{Yaghoubzadeh2015-ne,Yaghoubzadeh2016-re}. The Time Board is an important piece of incremental \dm, according to the authors, because not only does it maintain a history of the ongoing dialogue, future events (e.g., decisions) are also posted and coordinated. Events that have been initiated, for example the system begins an utterance that the \nlg\ is currently constructing and \tts\ is uttering, can clearly show that they are not yet complete, so a new event that needs to interrupt the ongoing event can produce natural behavior (e.g., saying \emph{um} or \emph{oops}, or \emph{sorry}). 

Going beyond rule-based incremental \dm, \citet{Selfridge2012-vo} introduced a first step towards an incremental \textsc{pomdp}-based system. They proposed an incremental interaction manager (\textsc{iim}) to mediate communication between an incremental \asr{} \ and a partially-observable \dm. The \textsc{iim} worked by evaluating potential \dm\ decisions by applying incremental \asr{} \ output to temporary instances of the \dm, allowing the system to maintain multiple \dm{} s across time and prune away \dm{} s that are unlikely to advance the dialogue. This enables the partially observable \dm\ to work with incremental \asr{} \ n-best lists, but the work demonstrated in \citet{Selfridge2012-ve} has regrettably not been pursued further. 

\begin{figure}
  \center
  \includegraphics[width=0.6\textwidth]{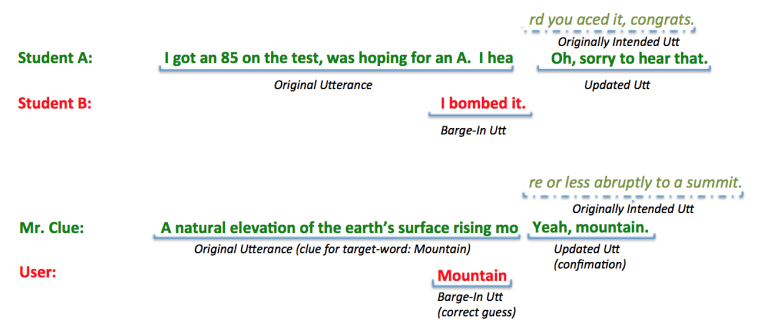}
  \caption{From \citet{Pincus2017-oh}, an example of human-human and game intelligent update dialogues with barge-in.}
  \label{fig:bargein}
\end{figure}

A \emph{barge-in} is when person A begins speaking, then person B attempts to take the floor while person A is still speaking. While often rude, this is common in interactive game scenarios, and it is important for a system that needs to have the ability to stop talking when a human barges in because timing is critical. \citet{Selfridge2013-dg} modeled a simple method for detecting barge-ins, and \citet{Pincus2017-oh} brought together multiple aspects of incremental dialogue in a word-game task that required fast-paced dialogue where barge-in was required. See Figure~\ref{fig:bargein} for an example. The system had an incremental \asr{} \ and learned a policy of when it should handle interruptions made while the system was speaking, and learning when to initiate barge ins. Though the focus was on barge-ins, there are some useful take-always from this work: first, that people often overlap in speech. Second, systems should be ready to yield the floor when they are barged-in on, and they should have the ability to barge in on a human's ongoing speech if there are appropriate stakes involved (e.g., a system needs to inform a human of an impending problem in a nuclear facility). None of these would be possible without incremental processing, and this work shows that timing is an important aspect of the kind of policy a \dm\ needs to learn about. 

\citet{Manuvinakurike2017-lu} also looked at incremental dialogue policy learning in a fast-paced game scenario where the user was presented with multiple images and needed to produce a referring expression to that object; the system was tasked with identifying which object the user was referring to. The system could highlight the image that it determined was being referred and say \emph{got it} or it could suggest that the system and user move onto the next set of images (e.g., ``let's move onto the next one") because it is unlikely to be able to refer to the correct one given the user's utterance. The learned policy was to either \texttt{wait} (i.e., let the user continue speaking), \texttt{As-I} (i.e., the system selects what it thinks the referred object is), or \texttt{As-S} (i.e., skip to the next set of images). The policy was incremental in that it had to learn at each word increment which action to take. The system and user earned points for identifying images quickly, but it lost points if it referred to an image incorrectly. The system, therefore, had to learn when to wait, select the image, or determine that it was better to move on. The evaluation showed that the learned policy worked better than the hand-coded policy in that it enabled more correctly identified images within a shorter amount of time. Like \citet{Pincus2017-oh}, the focus of this policy revolves around timing of simple actions rather than complex actions, indicating that the purpose of an incremental \dm\ should include handling timing decisions. 

Incremental \dm\ in a multi-party \hri\ setting was reported in \cite{Kennington2014-if}, that used the \iu\ framework, used an independent OpenDial \citep{Lison2016-jk} \dm\ for every human that it detected in a game setting. Overall, the \dm\ worked effectively, but it only controlled the dialogue interaction, not robot actions. 


\begin{figure}
  \center
  \includegraphics[width=0.6\textwidth]{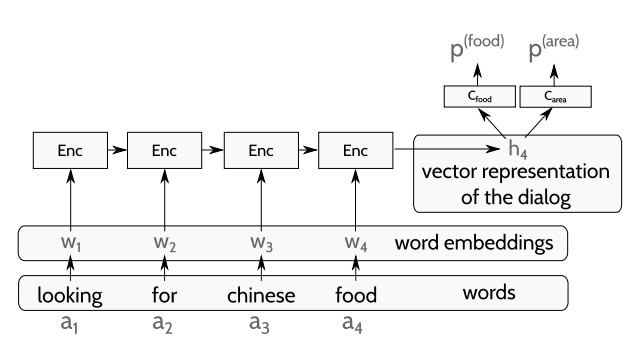}
  \caption{From \citet{Zilka2015-ul}, a schematic of a LSTM-based dialogue state tracker. }
  \label{fig:lstm-dst}
\end{figure}

Approaches to incremental dialogue state tracking have also been developed. \citet{Zilka2015-ul} introduced LecTrack, a word-level recurrent neural network state tracker model evaluated on DSTC2 data. The recurrent neural network they used as a long short-term memory (\textsc{LSTM}) because it is a kind of neural network that can be modeled to work at the word level and maintain its internal state (i.e., update-incremental) at each word increment (see Figure~\ref{fig:lstm-dst}). Their evaluations on a subset DTSC2 dataset showed as being on-par with state-of-the-art non-incremental state trackers. There also exists various approaches to dialogue state tracking based on autoregressive LMs \citep{Feng2023-tr,Hudecek2023-aj}, which rely on instruction-tuned LLMs to extract slot-value pairs from the dialogue history.

\section{Discussion}




One of the primary challenges of incremental \sds\ in \hri\ settings is handling uncertainty including sensory uncertainty and uncertainty that is inherent when communicating with humans. Certainly, all systems are required to handle uncertainty, but the problem is more acute with incremental, multimodal systems in \hri\ settings because they are tasked with acting on incomplete information that could be forthcoming at a later point. 

Evaluating \dm\ is challenging in general because a proper evaluation usually amounts to a fully-working \sds\ with human evaluation, but the \dm\ could be working perfectly while the \asr\ or the \nlg\ modules are not working properly for the task, resulting in poor evaluations from the humans. Offline evaluation is difficult for two reasons: while other modules like \asr, \nlu, and even \nlg\ can be evaluated with offline benchmarks, there isn't a clear offline evaluation for \dm, though the dialogue state tracking challenge is one attempt to address this. The second difficulty is that there is not a dataset that is annotated for \dm\ at an incremental level. It is therefore unknown whether a \dm\ \emph{should} make a decision at a specific point while a user is speaking, or how to handle errors in decisions when they are in the process of being articulated either in speech or a robotic action. The work on incremental \dm\ explained above \citep{Bus2011-es,Yaghoubzadeh2015-ne} show methods that attempt to address these challenges, but not with properly annotated incremental data.

\subsection{Desiderata}

To address these challenges, we offer here desiderata for incremental \dm\ on robotic platforms: 

\begin{itemize}
    \item \emph{Incremental \dm\ is responsible for timing}: knowing not just \emph{what} decision to make, but \emph{when} to make that decision are both important in fast-paced, incremental settings, particularly on robotic, embodied platforms where additional modalities play a role in understanding and interaction between user and system. \citet{DeVault2009-mq} explored how learning when to respond to incremental results affects task success, and \citet{Kennington2016-sa} used a rule-based \dm\ to make timing decisions on when to settle on a final decision on which action to take. Recent work relevant to \hri\ tasks in this area include \citet{Zhang2025-jv} that used a full-duplex \dm\ (i.e., the \asr\ was always producing input) based on Voice Activity Detection to help determine if the system should wait or act (AudioPaLM can likewise "speak and listen", which is a step in the right direction \citep{Rubenstein2023-jw}, and \citet{Yaghoubzadeh2015-ne}'s model of \dm\ that used a Time Board is a likely good place to start. 
   
    \item \emph{Incremental \dm\ needs to act on incomplete information}: When humans interact with each other, there are often backchannels (e.g., nodding) that signal understanding, or as someone is speaking a listener can signal understanding by taking an action. For example, if a speaker makes a request \emph{can you hand me the green book on the left?} the listener can already be turning and reaching for a green book before the utterance is complete. A robot that interacts with a person where the task involves handling objects should act in a similar way; for example, reaching for an object or driving towards a destination. If indeed the system made the wrong decision about which object to pursue, then the robot can change its course, but it's important that the robot act as soon as it has enough information to act, even if that act might be incorrect; the movement signals to the user that the robot is in the process of understanding. 
    
    \item \emph{Incremental \dm\ needs to make fast, small decisions concurrently}: Traditional \dm s often take in all information from the user during their turn then make a high-level decision once which can then potentially inform multiple modules like \nlg\ to speak and a robot arm to move. An incremental \dm, in contrast, needs to make smaller decisions that may lead to a final outcome, but the outcome may not yet be known. This is similar in principle to acting on incomplete information, but the nature of the actions is more fine-grained. For example, the \dm\ may know that the user wants a robot to fetch an object in the kitchen, and though the robot doesn't know which object, it makes a smaller decision to move to the kitchen, and by the time it arrives in the kitchen it knows more about the specifics of the object that it is requested to retrieve. 
    
    \item \emph{Incremental LLMs}: Transformer LLM architectures generate output incrementally and some aspects of input are incremental, but they are not completely update-incremental. Using them in a restart-incremental manner is computationally expensive, but recent work has shown that minor changes to the model can improve incremental metrics and reduce computational overhead \citep{Kahardipraja2021-fv}. LLMs are being used in many ways in robotics (see, for example, \citet{Singh2023-ea} that uses LLMs for robot action planning), but work needs to be done for incremental processing on LLMs in \hri\ settings.
\end{itemize}

\paragraph{Recommendations} There is a lack of incremental datasets. Most datasets can be used for incremental training and evaluation for some modules (e.g., \asr\ or \nlg), but \nlu\ and \dm\ modules that produce incremental output that is on a different level of granularity than the word level, so it is unclear from \nlu\ datasets as to \emph{when} a slot should be filled or \emph{when} the \dm\ should make a decision. Efforts towards a dataset that has incremental annotations would be very beneficial to research in the setting of dialogue with robots. Models may or may not need to be trained incrementally, but evaluation metrics should be on the incremental level. 

\section{Conclusion} 

In this article, we reviewed literature relating to incremental dialogue management motivated by the need for incremental dialogue management in robotic platforms. We showed that there is ample work in incremental processing, but very little in incremental dialogue management itself. The review resulted in several key desiderata for incremental dialogue management, particularly needed in spoken dialogue-enabled human robot interaction. 

Clearly, a decision-making module is a critical component in a robot that can interact with people using spoken dialogue. Taken together, this review in conjunction with other recent review work from \citet{Reimann2023-ui} and \citet{Lison2023-mu} are useful for robotics researchers who are interested in designing effective dialogue strategies between robots and humans. 


\bibliography{refs,paperpile}

\end{document}